\documentclass[runningheads]{llncs}
\usepackage[T1]{fontenc}
\usepackage{graphicx}
\usepackage{xcolor}
\usepackage{url}

\usepackage{hyperref}

\hypersetup{
    colorlinks=true,
    urlcolor=blue,    
    linkcolor=black,
    citecolor=black,
    anchorcolor=black
}

\usepackage[numbers]{natbib}
\usepackage{booktabs}
\usepackage{multirow}
\usepackage{makecell}
\usepackage{amssymb}
\usepackage{adjustbox}
\usepackage{float}
\usepackage{subcaption}
\usepackage{amsmath}
\usepackage{bbding}

\usepackage{marvosym}
\begin{document}
\title{Towards Solving Polynomial-Objective Integer Programming with Hypergraph Neural Networks}
\titlerunning{Towards Solving Polynomial-Objective Integer Programming with HNN}
%
\author{Minshuo Li\inst{1,2} \and
Yaoxin Wu\inst{2,3(\mbox{\Letter})} \and
Pavel Troubil\inst{4} \and
Yingqian Zhang\inst{2,3} \and
Wim P.M. Nuijten\inst{1,2}}
\authorrunning{M. Li et al.}
%
\institute{Department of Mathematics and Computer Science, Eindhoven University of Technology, Eindhoven, the Netherlands \\
\email{\{m.li7, W.P.M.Nuijten\}@tue.nl}
\and
Eindhoven Artificial Intelligence Systems Institute, Eindhoven University of Technology, Eindhoven, the Netherlands \\
\email{\{y.wu2, yqzhang\}@tue.nl}
\and
Department of Industrial Engineering and Innovation Sciences, Eindhoven University of Technology, Eindhoven, the Netherlands \\
\and
Delmia R\&D, Dassault Systèmes, ’s-Hertogenbosch, the Netherlands \\
\email{pavel.troubil@3ds.com}
}
\maketitle              
\begin{abstract}
Complex real-world optimization problems often involve both discrete decisions and nonlinear relationships between variables.
Many such problems can be modeled as polynomial-objective integer programs, encompassing cases with quadratic and higher-degree variable interactions. Nonlinearity makes them more challenging than their linear counterparts. In this paper, we propose a hypergraph neural network (HNN) based method to solve polynomial-objective integer programming (POIP). Besides presenting a high-degree-term-aware hypergraph representation to capture both high-degree information and variable-constraint interdependencies, we also propose a hypergraph neural network, which integrates convolution between variables and high-degree terms alongside convolution between variables and constraints, to predict solution values. Finally, a search process initialized from the predicted solutions is performed to further refine the results. Comprehensive experiments across a range of benchmarks demonstrate that our method consistently outperforms both existing learning-based approaches and state-of-the-art solvers, delivering superior solution quality with favorable efficiency. Note that our experiments involve both polynomial objectives and constraints, demonstrating our HNN’s versatility for general POIP problems and highlighting its advancement over the existing literature.

\keywords{Polynomial-objective integer programming  \and Hypergraph neural network \and Nonlinear integer programming.}
\end{abstract}
\section{Introduction}
\label{sec:intro}

Integer programming has been widely applied to real-world applications involving discrete decisions, such as photolithography scheduling~\cite{deenen2023scheduling} and supply chain optimization~\cite{bai2011biofuel}. Many integer programming problems are NP-hard, meaning that the computational time required to reach optimality grows exponentially with problem size. 
In particular, nonlinear integer programming (NLIP) frequently arises due to physical laws, statistical measures, nonlinear regression, and other complex relationships~\cite{ahmadi2016some}. The nonlinearity makes these problems even more challenging to solve than their linear counterparts, highlighting the need for efficient solution methods that go beyond traditional techniques.

Over the past few decades, different algorithms have been proposed to address the challenges of NLIP, typically following two main approaches. Local approaches rely on gradient information to find locally optimal solutions~\cite{bazaraa2006nonlinear}, but often struggle with complex structures containing multiple local optima. Global approaches follow a divide-and-conquer strategy, partitioning the solution space and then searching within each partition to identify the optimal solution~\cite{smith1999symbolic, kesavan2004outer},  
This often incurs prohibitive computational times for instances with high nonlinearity or intricate constraint structures. Furthermore, algorithms for these approaches are typically closed-source or tailored to specific NLIP problems, which restricts their broader application~\cite{kronqvist2019review}. These limitations motivate the exploration of alternative paradigms.

A promising alternative paradigm is machine learning, which has recently made major advances in integer linear programming (ILP) The advances include learning better policies within specific solvers, such as branching~\cite{gasse2019exact, nair2021solvingmixedintegerprograms, Maudet_Danoy_2025} and presolving~\cite{ICLR2024_1e6e0c2e, 10971949}, as well as learning general guidance for ILP solvers, such as solution prediction~\cite{ding_accelerating_2020, geng2025differentiable} and neighborhood selection~\cite{DBLP:conf/iclr/HanYCZZW0L23, ye2023gnn}. On the one hand, due to the fundamental differences between linear and nonlinear formulations, these methods are not directly applicable to NLIP. On the other hand, research on learning-based methods for NLIP remains relatively limited, with only a handful of works~\cite{ghaddar2023learning, ferber2023surco}. These methods are mainly built on specific problem structures or algorithms, thus restricting their broader applicability. Therefore, there is a need for learning-based methods that can effectively address a wide range of NLIP problems and work seamlessly across different solvers.

To address these limitations, this paper aims to push the frontier of learning-for-NLIP towards solving polynomial-objective integer programming (POIP), an important subclass of NLIP.
By Taylor's formula~\cite{rudin1987real}, POIP captures many practical nonlinearities and is representative of NLIP challenges that cannot be efficiently solved by current solvers. More  specifically, we innovate a learning-based approach to NLIP and propose a hypergraph neural network (HNN) based model that predicts variable values in optimal solutions based on a hypergraph representation of problem instances. The predicted solution serves as an effective initial solution, which can be further refined by any solver or complementary search algorithm. Our major contributions are summarized as follows.

\begin{itemize}
    \item A high-degree-term-aware hypergraph representation for general POIP, which encodes interactions among variables within high-degree terms and relations between variables and constraints to reflect problem structure.
    \item A hypergraph neural network that learns the mapping between instances' hypergraph representations and their corresponding optimal solutions. It applies a convolution across variables and high-degree terms to capture variable representations,
    together with a convolution between variables and constraints to further involve variable-constraint interdependencies. 
    \item Experimental results on diverse benchmark datasets demonstrate that our method significantly improves the efficiency of Gurobi and SCIP on quadratic and quintic integer programming problems, consistently outperforming existing learning-based approaches.
\end{itemize}

\section{Related Work}
\label{sec: related_work}

\subsection{Learning-based Methods for ILP}
\label{sec: related_work-ILP}
Learning-based methods for ILP can be broadly categorized into two classes~\cite{zhang2023survey}. The first class concerns learning key policies within solvers. Among these, most studies learn variable selection~\cite{gasse2019exact, 10.5555/3495724.3497242, nair2021solvingmixedintegerprograms, feng2025sorrel, li2025towards} and node selection~\cite{10.5555/3600270.3602589, Maudet_Danoy_2025} in branch and bound (B\&B) tree search. Some works learn other policies including cutting plane selection~\cite{deza2023machine, 10.1109/TPAMI.2024.3432716}, primal heuristic selection~\cite{chmiela2021learning}, and presolving settings~\cite{ICLR2024_1e6e0c2e, 10971949}. The second class focuses on learning general policies that are applicable across different solvers. One approach involves predicting solutions, either as initial solution values for further refinement~\cite{ding_accelerating_2020, pmlr-v235-huang24f} or as direct feasible solutions~\cite{geng2025differentiable, liu2025apollomilp}. Another research direction involves learning strategies in general heuristics, including neighborhood selection strategies for large neighborhood search~\cite{liu2022learning, ye2023gnn, DBLP:conf/iclr/HanYCZZW0L23, wu2021learning} and search strategies for diving heuristics~\cite{nair2021solvingmixedintegerprograms}.

\subsection{Learning-based Methods for NLIP}
\label{sec: related_work-NLIP}

Learning-based methods for NLIP remain underexplored in the literature. A few noteworthy contributions have developed learning-for-NLIP methods for specific problems. 
Bonami et al.~\cite{bonami2022classifier} train a classifier to decide whether linearizing quadratic integer programs leads to better solver performance. 
Ferber et al.~\cite{ferber2023surco} propose to learn surrogate linear objective functions for nonlinear programs with linear constraints. Tang et al.~\cite{tang2024learning} introduce differentiable correction layers for end-to-end learning on parametric nonlinear programming with fixed problem structure. Chen et al.~\cite{chen2025expressive} and Wu et al.~\cite{wu2025on} prove the theoretical expressive power of graph neural networks for quadratic terms. While these works represent valuable progress, they are all tailored to specific problem settings and do not generalize to polynomial-objective integer programming (POIP).

Two recent studies are more directly related to our work. Xiong et al. ~\cite{xiong2024neuralqp} develop a graph neural network to predict solutions for quadratic programming. However, both their graph representation and neural network are restricted to quadratic terms, whereas our framework is designed to handle POIP with arbitrary degrees. Ghaddar et al.~\cite{ghaddar2023learning} propose quantile regression to learn instance-specific branching rules inside a closed-source solver RAPOSa~\cite{gonzalez2023computational} for solving polynomial programming. This approach requires access to internal solver modifications and is limited to a specific solver. Instead, our approach predicts solutions and serves as an external module to provide initial solution values for any solver or search algorithm, without the requirement of internal changes.

The literature review highlights the limitations of current learning-based methods for NLIP. Our work pushes the frontier of this domain by proposing a hypergraph neural network framework, which addresses integer programming with high-degree objective terms and integrates seamlessly with existing solvers.

\section{Preliminaries}
\label{sec: pre}

\subsection{Definition of Polynomial-Objective Integer Programming}
\label{sec: pre-iphd}

Without loss of generality, polynomial-objective integer programming (POIP) refers to a class of optimization problems that maximizes a polynomial objective function defined over a set of integer variables while satisfying a set of constraints. A generic POIP formulation can be written as:
\begin{align} \label{problem_defination}
    \max_{x} \quad & f(x_1, x_2, \cdots, x_n), \\
    \text{s.t.} \quad & \sum_{i=1,\cdots,n} a_{ij} x_i \leq b_j, \quad j = 1,2,\ldots,m, \\
    & l_i \leq x_i \leq u_i, x_i \in \mathbb{Z}, \quad i = 1,2,\ldots,n,
\end{align}
where $n$ is the number of variables and $m$ is the number of constraints; $f$ is a polynomial expression over the variables $x_1, x_2, \cdots, x_n$; $a_{ij}$ and $b_j$ are, respectively, the coefficient of the variable $x_i$ and the right-hand side scalar in the $j$-th constraint; $l_i$ and $u_i$ are the lower and upper bounds for $x_i$.

\subsection{Graph and Hypergraph Neural Networks} 
\label{sec: preliminar-hypergraph}
A graph is defined as $G=(V,E)$, where $V$ is the set of vertices and $E \subseteq V \times V$ is the set of pairwise edges. Graph Neural Networks (GNNs) learn node or edge representations by iteratively aggregating and transforming information from local neighborhoods~\cite{wu2020comprehensive}. Owing to their ability to exploit graph structure, GNNs have been applied to a variety of IP problems that have natural graph representations~\cite{SMIT2025106914, bengio2021machine, cappart2023combinatorial,wu2022graph}.

A hypergraph extends a graph by allowing edges, called hyperedges, to connect any number of vertices. This enables the modeling of higher-order relationships that cannot be captured by pairwise edges alone. Hypergraph Neural Networks (HNNs) extend GNN principles to learning representations of hypergraphs. They have shown promising performance across diverse hypergraph-based applications~\cite{kim2024survey}. A representative framework is UniGNN~\cite{ijcai21-UniGNN}, which introduced a two-stage message passing mechanism:
\begin{equation} \label{unignn}
    \text{(UniGNN)} 
    \begin{cases} 
    h_{\epsilon} = \phi_{1} (\{ h_v \}_{v \in \mathcal{N}_\epsilon}), \\ 
    \bar{h}_v = \phi_{2} (h_v, \{h_\epsilon\}_{\epsilon \in \mathcal{N}_{v}}), 
    \end{cases}
\end{equation}

\noindent where $h_{\epsilon}$ and $h_v$ denote the representations for hyperedge $\epsilon$ and vertex $v$, respectively. The sets $\mathcal{N}_\epsilon$ and $\mathcal{N}_v$ represent all the vertices contained in the hyperedge $\epsilon$ and all the hyperedges containing the vertex $v$, respectively. The functions $\phi_{1}$ and $\phi_{2}$ are permutation-invariant aggregation functions, and $\bar{h}_v$ represents the updated representation of vertex $v$. The two-stage message passing mechanism in Eq.~\ref{unignn} can be iteratively executed to progressively refine the vertex and hyperedge embeddings.

\subsection{Graph Representations for Integer Programming}
\label{sec: pre-hnn}
Graph representations are commonly used to transform integer programming (IP) instances into structures suitable for graph neural network processing. Gasse et al.~\cite{gasse2019exact} introduce a bipartite graph, in which one set of vertices represents variables and the other set represents constraints, with edges encoding variable-constraint incidences, i.e., a variable appearing in a constraint. Ding et al.~\cite{ding_accelerating_2020} extend the representation to a tripartite graph by adding a vertex to represent the objective function, thereby enriching the structural information. 
Subsequent graph representations of IP are often built on bipartite and tripartite graphs, because of their effectiveness and simplicity.

While effectively capturing variable-constraint relationships, current graph representations are restricted to pairwise interactions and thus struggle to model nonlinear or higher-order structures that frequently arise in practical IP problems. To overcome this limitation, Heydaribeni et al.~\cite{heydaribeni2024distributed} use hyperedges to connect variables appearing in the same constraint, and Xiong et al.~\cite{xiong2024neuralqp} employ hyperedges to represent quadratic terms involving both variables and constraints. Both graph representations remain limited, for example to IP with linear or quadratic terms, and cannot generalize to IP instances with arbitrary high-degree terms. In this paper, we address this gap by proposing a hypergraph neural network tailored for learning over graph representations of POIP.

\section{Methodology}
\label{sec: method}

\begin{figure}[t]
  \centering
  \includegraphics[width=1\linewidth]{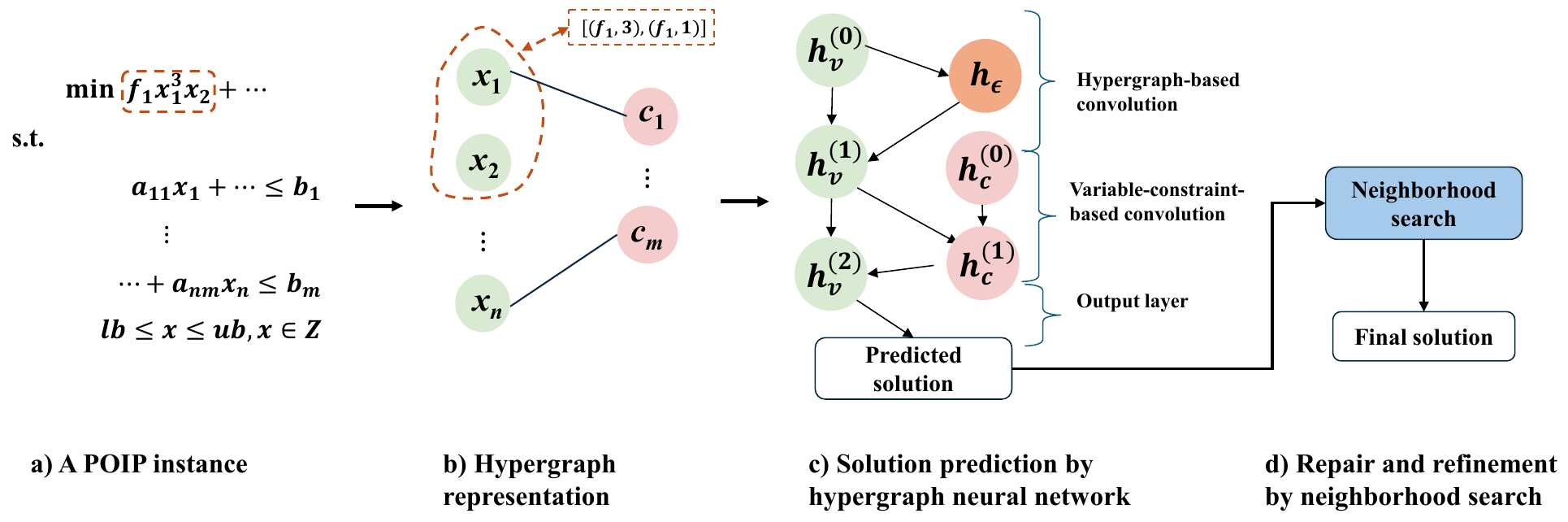}
  \caption {\textbf{The framework of the proposed method.} For a POIP instance shown in a), our method first transforms it into a hypergraph shown in b), where orange dashed circles denote hyperedges representing high-degree terms. Their raw features (term coefficients and variable degrees) are illustrated in the orange dashed boxes. This hypergraph is then processed by a hypergraph neural network shown in c) for representation learning and solution prediction, where $h_{\epsilon}$, $h_v^{(i)}$, and $h_c^{(i)}$ represent hyperedge embeddings, variable embeddings after the $i$-th update, and constraint embeddings after the $i$-th update, respectively. Finally, a neighborhood-search based repair-and-refinement process, shown in d), turns the predicted results into a high-quality feasible solution.}
  \label{fig:framework}
\end{figure}

This section presents our hypergraph neural network framework for tackling POIP problems, including hypergraph representation, solution prediction via hypergraph neural network, and solution repair and refinement. The overview of our framework is illustrated in Figure~\ref{fig:framework} and detailed below.

\subsection{High-Degree-Term-Aware Hypergraph Representation}
\label{sec: method-representatoin}

Representing general POIP poses two unique challenges: (i) high-degree terms induce multi-variable interactions that cannot be captured by standard pairwise connections~\cite{gasse2019exact}, and (ii) the satisfaction of constraints depends intricately on variable assignments, forming the essential variable-constraint relationship. To address these challenges, we encode a POIP instance as a hypergraph. 

Formally, our hypergraph is defined as $\mathcal{G}=(\mathcal{V}, \mathcal{C}, \mathcal{H}, \mathcal{E})$, where $\mathcal{V}$ denotes variable vertices, $\mathcal{C}$ denotes constraint vertices, $\mathcal{H}$ denotes hyperedges for high-degree objective terms, and $\mathcal{E}$ are edges encoding variable–constraint incidence. Specifically, each variable in a POIP instance is represented by a vertex $v \in \mathcal{V}$, and each constraint is represented by a vertex $c \in \mathcal{C}$.  
For every high-degree term $c_{\alpha} \prod_{i^{\prime}\in\{1,\cdots,|\mathcal{V}|\}} x_{i^{\prime}}^{\alpha_{i^{\prime}}}$ appearing in the objective function, we create a hyperedge $\epsilon \in \mathcal{H}$ that connects all variables contained in the term. A raw feature $[\omega_{v_{i^{\prime}}\epsilon}=(c_{\alpha},\alpha_{i^{\prime}})]_{v_{i^{\prime}}\in\mathcal{N}_{\epsilon}}$ is assigned to the hyperedge $\epsilon$, where $\mathcal{N}_{\epsilon}$ denotes the variables contained in $\epsilon$.
For variable-constraint relationships, we add an edge $e_{vc} \in \mathcal{E}$ between variable vertex $v$ and constraint vertex $c$ whenever the corresponding variable appears with a nonzero coefficient in the corresponding constraint. The associated coefficient is assigned to the edge's feature, ensuring that the numerical dependency between variable and constraint is preserved. 

The high-degree-term-aware hypergraph representation integrates both structural and parametric information of a POIP instance, 
and provides a foundation for the hypergraph neural network introduced in the next subsection. A complete specification of the raw features is provided in Appendix A.1.

For example, Figure~\ref{fig:framework}(b) illustrates the high-degree-term-aware hypergraph representation for the POIP instance in Figure~\ref{fig:framework}(a). Variable vertices (left) and constraint vertices (right) represent variables and constraints separately; edges (straight lines) represent variable-constraint relationships, connecting a variable to each constraint in which it has a nonzero coefficient (e.g., $x_1$ is connected to $c_1$);
hyperedges (circles) capture the relationships of variables in high-degree terms in the objective. For the term $f_1 x_1^3 x_2$, the hyperedge feature set is $\{(f_1,3),(f_1,1)\}$, associating the shared coefficient with each variable’s exponent.

\subsection{Solution Prediction via Hypergraph Neural Network}
\label{sec: method-hnn}

Based on the hypergraph representation of POIP instances, our neural network is designed to capture two complementary types of relationships: (i) higher-order interactions among variables induced by high-degree terms, and (ii) interdependencies between variables and constraints. To achieve this, we draw on concepts from Hypergraph Neural Networks (HNNs)~\cite{kim2024survey}, which support message passing between vertices and hyperedges. Building on this framework, we first introduce a hyperedge-based convolution that aggregates information from hyperedges into variable embeddings, enabling the model to represent higher-order structures. We then complement this with a variable–constraint convolution that propagates information along standard edges, thereby modeling the dependencies between variables and constraints.
Finally, the variable embeddings are passed through an output layer to generate the predictions of variable values in high-quality solutions. The architecture of our model is illustrated in Figure~\ref{fig:framework}(c).

\subsubsection{Hyperedge-based Convolution}
\label{sec: method-hnn-hyper}
Our HNN begins with a hyperedge-based convolution applied to hyperedges and their associated variable vertices in order to effectively extract higher-order information arising from the high-degree terms. Inspired by UniGNN~\cite{ijcai21-UniGNN} (see Section~\ref{sec: preliminar-hypergraph}), we formulate the convolution as presented in Eq.~\ref{hyper-conv1} and Eq.~\ref{hyper-conv2}. Each iteration updates hyperedge embeddings by aggregating incident variable information (Eq.~\ref{hyper-conv1}), then updates variable embeddings by aggregating the connected hyperedge information (Eq.~\ref{hyper-conv2}). In this way, information from high-degree terms is integrated into the embeddings of the variables.
\begin{equation} \label{hyper-conv1}
    h_{\epsilon} \leftarrow \sum_{v \in \mathcal{N}_{\epsilon}} h_v h_{v\epsilon}, \forall \epsilon\in\mathcal{H},
\end{equation}
\begin{equation} \label{hyper-conv2}
    h_v \leftarrow \phi_\mathcal{H}(h_v, \text{mean}(\{h_\epsilon h_{v\epsilon}\}_{\epsilon \in \mathcal{N}_{v}^H}))+h_v, \forall v\in\mathcal{V},
\end{equation}
Formally, $h_v$, and $h_{v\epsilon}$ denote the embeddings for variable vertex $v$ and  feature vector $\omega_{v\epsilon}$ (see Section~\ref{sec: method-representatoin}), respectively. These embeddings are initialized from raw features using two-layer MLPs, and the subsequent iterations adopt the embeddings from the previous iteration (where Eq.~\ref{hyper-conv1} and Eq.~\ref{hyper-conv2} are applied) as input. $h_{\epsilon}$ denotes the embeddings for hyperedge $\epsilon$, and both its initialization and subsequent updates are calculated by Eq.~\ref{hyper-conv1}.  $\mathcal{N}_{v}^H$ denotes the set of hyperedges containing the variable $v$; $\mathcal{N}_{\epsilon}$ denotes the set of variable vertices contained in the hyperedge $\epsilon$. The function $\phi_\mathcal{H}$ is parameterized by another two-layer MLP activated by LeakyReLU.
The hyperedge-based convolution (i.e., Eq.~\ref{hyper-conv1} and Eq.~\ref{hyper-conv2}) is repeated for $L$ iterations to fully capture information from high-degree terms.


\subsubsection{Variable-Constraint-based Convolution}
\label{sec: method-hnn-bi}
After the hyperedge-based convolution embeds higher-order relationships into variable vertices, the model still needs to account for variable-constraint interdependencies. To this end, we apply a variable-constraint-based convolution that explicitly processes message passing along the edges connecting variable and constraint vertices.

This convolution operates through bidirectional message passing: variable embeddings (that already aggregate higher-order information from the hyperedge-based convolution) are first propagated to constraint vertices (Eq.~\ref{bi-conv1}), and then, the updated constraint representations are passed back to the variable vertices (Eq.~\ref{bi-conv2}). Through the two-step convolutions, the variable embeddings are enriched with information of the constraints they are required to satisfy.

\begin{equation} \label{bi-conv1}
    h_c \leftarrow f_\mathcal{C}(h_c, \sum_{v\in \mathcal{N}_c} \phi_{\mathcal{C}}(h_c, h_v, h_{vc}))+h_c, \forall c \in \mathcal{C},
\end{equation}
\begin{equation} \label{bi-conv2}
    h_v \leftarrow f_\mathcal{V}(h_v, \sum_{c\in \mathcal{N}_v^C} \phi_{\mathcal{V}}(h_c, h_v, h_{vc}))+h_v, \forall v \in \mathcal{V},
\end{equation}
Formally, $h_c$ and $h_{vc}$ denote the embeddings for the constraint vertex $c$ and the edge $e_{vc}$ connecting $v$ and $c$, separately, and they are initialized from the raw features of constraint vertices and edges via two-layer MLPs. $h_v$ denotes the embedding for variable vertex $v$ and is passed from the hyperedge-based convolution. The set $\mathcal{N}_c$ contains all variable vertices connected to constraint $c$, while $\mathcal{N}_v^C$ contains all constraint vertices connected to variable $v$. Finally, $\phi_{\mathcal{C}}$, $\phi_{\mathcal{V}}$, $f_\mathcal{C}$, and $f_\mathcal{V}$ are implemented as two-layer MLPs activated by LeakyReLU. 
The variable-constraint-based convolution is executed once.

\subsubsection{Solution Prediction and Refinement}
\label{sec: method-search}
After hyperedge-based and variable-constraint-based convolutions, the variable embeddings involve both high-order interactions from high-degree terms and the interdependencies between variables and constraints. We finally feed these embeddings into a two-layer MLP, which outputs a scalar value for each variable as the solution prediction.

Without loss of generality, we focus on POIP with binary variables. This is reasonable because any bounded integer variable can be binarized, enabling a bounded POIP to be reformulated as an equivalent binary POIP~\cite{dantzig2016linear}. The entire HNN is trained in a supervised manner using the binary cross-entropy loss:
$$\mathcal{L_{\text{BCE}}} = -\frac{1}{N}\sum_{i=1}^{N} [y_i \log(\sigma(\hat{y}_i)) + (1-y_i) \log(1-\sigma(\hat{y}_i))],$$ 
where $N$ is the number of logits from the HNN, $y \in \{0,1\}^N$ and $\hat{y} \in \mathbb{R}^N$ represent the ground truth labels and the predicted logits, and $\sigma(\cdot)$ is the sigmoid function.

The predictions produced by our model provide initial solution values for downstream algorithms and solvers. To make use of them, we adopt the parallel neighborhood optimization framework in~\cite{ye2023gnn, xiong2024neuralqp} that embeds an off-the-shelf solver to refine the predicted solution during inference. Specifically, the framework employs adaptive large neighborhood search. It first repairs the predictions into feasible solutions by fixing promising variables to their predicted values while allowing the remaining variables to be reoptimized by a solver, and then further refines these feasible solutions through additional neighborhood search to achieve better objective values. We refer interested readers to~\cite{ye2023gnn, xiong2024neuralqp} for more details.

\section{Experimental Results}

\label{sec: exp}
In this section, we conduct comprehensive experiments to demonstrate the effectiveness of our proposed method in solving POIP instances. 
Section~\ref{sec:exp-setup} describes the experimental setup. Section~\ref{sec: exp-results} reports comparative results on quadratic and quintic POIP benchmarks. Section~\ref{sec: exp-generalization} investigates generalization to public datasets and quadratically constrained benchmarks. Finally, Section~\ref{sec: exp-ablation} presents an ablation study on the model architecture and high-degree term handling.

\subsection{Setup}
\label{sec:exp-setup}

\paragraph{Benchmark instances}
Our experiments were conducted on three POIP benchmark problem instances. The first one is a synthetic quadratic integer programming (QIP) benchmark introduced by Xiong et al.~\cite{xiong2024neuralqp}, derived from the Quadratic Multiple Knapsack Problem (QMKP). It contains instances at five scales (Mini, 1k, 2k, 5k, 10k), where the first three scales are used for training and all except Mini are included in testing. The second benchmark contains hard quadratic instances from the public library QPLIB~\cite{furini2019qplib}, evaluated also by Xiong et al.~\cite{xiong2024neuralqp}. Third, we introduce a new synthetic quintic integer programming benchmark based on the Capacitated Facility Location Problem with Traffic Congestion (CFLPTC). This dataset includes five instance scales (50$\times$10, 50$\times$20, 150$\times$30, 200$\times$30, 500$\times$100), with the first four scales used for training and the last three for testing. Detailed formulations and descriptions for all benchmarks are provided in Appendix B.

\paragraph{Baselines}
We compare our method against three recent learning-based methods tailored for quadratic programming (QP): (1) NeuralQP~\cite{xiong2024neuralqp} that introduces a hypergraph neural network to predict solutions for quadratically constrained QPs,  (2)  GNNQP~\cite{chen2025expressive} and (3) TriGNN~\cite{wu2025on}, both investigating the theoretical expressive power of graph neural networks for quadratic terms.  Similar to our HNN method, these models serve as solution predictors and are combined with exact solvers for repair and refinement (see Section~\ref{sec: method-search}). We adopt SCIP and Gurobi as the exact solvers in this experiment, and we also include them as standalone baselines to provide a comprehensive comparison.

\paragraph{Metrics}
We evaluate performance using the relative primal gap (in percentage), defined as $\mathrm{gap_{\%}}=|\mathrm{OBJ}-\mathrm{BKS}|/(|\mathrm{BKS}|+10^{-10})\times 100$, where $\mathrm{OBJ}$ is the objective value obtained by a method and $\mathrm{BKS}$ is the best-known solution of the instance. $\mathrm{BKS}$ values of QPLIB instances are publicly available, while those for synthetic datasets are the best objective value found across our experiments. Under the same time limit, a lower $\mathrm{gap_{\%}}$ indicates solutions closer to $\mathrm{BKS}$ and thus stronger performance. Wilcoxon signed-rank test at the 95\% confidence level is applied to assess statistical significance of the difference among methods.

\paragraph{Implementations}
All learning-based methods follow the same evaluation procedure. Each trained model first generates a solution prediction, which is then improved using the repair-and-refinement strategy (Section~\ref{sec: method-search}) under a fixed time limit. 
The time limits vary by benchmark and instance scale: 100, 600, 1800, and 3600 seconds for QMKP instances at scales 1k, 2k, 5k, and 10k, respectively; 100 seconds for QPLIB instances; and 60, 60, and 1000 seconds for CFLPTC instances at scales 150$\times$30, 200$\times$30, and 500$\times$100. The exact-solver baselines (SCIP and Gurobi) are given the same time budgets to solve each instance from scratch. Each method was run five times per instance. 

For inference, we matched training and testing benchmarks where possible. On the QMKP and CFLPTC datasets, models trained on the same benchmark with identical or smaller instance scales were used for testing. For QPLIB, the models were trained on QMKP-1k instances. 
More details on experimental implementations can be found in Appendix C.

\begin{table}[t]
  \centering
  \caption{Comparison on QMKP datasets in terms of mean and standard deviation of $\mathrm{gap_{\%}}$. The best results are highlighted in bold and $^{*}$ indicates the overall results are statistically different from the best one. ``--'' denotes that the corresponding test was omitted, as models were only evaluated on instances with identical or larger scales than the training instances, following the setting in~\cite{xiong2024neuralqp}.}
  \label{QKP_results}
  \setlength{\tabcolsep}{1pt}
  \fontsize{9}{11}\selectfont
  \renewcommand{\arraystretch}{0.8}
  \begin{tabular}{cc|ccccc|ccccc}
    \toprule
        \multirow{2}{*}{Method} & \multirow{2}{*}{Train} & \multicolumn{5}{c|}{Base solver: Gurobi} & \multicolumn{5}{c}{Base solver: SCIP} \\
        \cmidrule(lr){3-12}
        & & 1k & 2k & 5k & 10k & Overall & 1k & 2k & 5k & 10k & Overall \\
        \midrule
        \begin{tabular}[c]{@{}l@{}}Exact\\solver\end{tabular} & -- & $0.38$ & $0.23$ & $29.08$ & $26.69$ & 14.10$_{\pm 13.89}^{*}$ 
        & $5.15$ & $6.10$ & $35.73$ & $27.61$ & 18.65$_{\pm 13.38}^{*}$ \\
        \midrule
        \multirow{3}{*}{\begin{tabular}[c]{@{}l@{}}Neural\\QP\end{tabular}} & Mini & $0.35$ & $0.09$ & $0.03$ & $0.03$ & \multirow{3}{*}{0.11$_{\pm 0.13}$} 
        & 0.54 & 0.11 & 29.38 & 17.90 & \multirow{3}{*}{12.99$_{\pm 13.21}$}\\
         & 1k & $0.35$ & $0.09$ & $0.04$ & $0.04$ & 
         & 0.38 & 0.11 & 28.68 & 19.44 & \\
         & 2k & -- & $0.09$ & $0.05$ & $0.04$ & 
         & -- & \textbf{0.10} & 29.54 & 17.41 & \\
        \midrule
        \multirow{3}{*}{GNNQP} & Mini & $0.45$ & $0.09$ & $0.05$ & $0.03$ & \multirow{3}{*}{7.22$_{\pm 15.00}^{*}$} 
        & 0.54 & 0.11 & 29.38 & 17.90 & \multirow{3}{*}{16.08$_{\pm 15.52}$} \\
         & 1k & $0.49$ & $0.10$ & $0.06$ & $38.95$ & 
         & 0.52 & 0.11 & 29.48 & 38.09 & \\
         & 2k & -- & $0.09$ & $0.05$ & $39.05$ & 
         & -- & 0.11 & 29.53 & 31.12 & \\
         \midrule
        \multirow{3}{*}{TriGNN} & Mini & $0.43$ & $0.09$ & $0.04$ & $0.04$ & \multirow{3}{*}{0.13$_{\pm 0.17}$} 
        & 0.52 & 0.12 & 28.70 & \textbf{17.25} & \multirow{3}{*}{12.87$_{\pm 12.95}$} \\
         & 1k & $0.46$ & $0.09$ & \textbf{0.03} & $0.04$ & 
         & 0.47 & 0.10 & \textbf{28.63} & 18.77 & \\
         & 2k & -- & $0.10$ & $0.03$ & $0.04$ & 
         & -- & 0.12 & 28.66 & 18.26 & \\
         \midrule
        \multirow{3}{*}{Ours} & Mini & \textbf{0.16} & 0.08 & $0.03$  & \textbf{0.03} & \multirow{3}{*}{\textbf{0.09}$_{\pm 0.10}$} 
        & \textbf{0.18} & 0.10 & 28.67 & 17.93 & \multirow{3}{*}{\textbf{12.76}$_{\pm 13.21}$} \\
         & 1k & $0.31$ & $0.10$ & $0.05$  & $0.04$  & 
         & 0.33& 0.11& 29.33 & 16.79 & \\
         & 2k & -- & \textbf{0.07} & $0.05$  & $0.03$  & 
         & -- & 0.10 & 29.41 & 17.45 & \\
    \bottomrule
  \end{tabular}
\end{table}

\subsection{Performance on Polynomial-Objective Integer Programs}
\label{sec: exp-results}

We first evaluate all methods on synthetic benchmarks: the QMKP benchmark and the CFLPTC benchmark. Results are shown in Table~\ref{QKP_results} and Table~\ref{tab:cflptc_comparison}.

On QMKP, we compare our method against all learning-based baselines and the two exact solvers. Table~\ref{QKP_results} reports the mean and standard deviation of $\mathrm{gap_{\%}}$ across instance scales. It shows that the learning-based methods reach at least comparable solution qualities compared to exact solvers on 1k-scale instances and achieve better performance on larger-scale instances, highlighting the promise of learning-based approaches. Our method achieves performance that is at least comparable to, and in several cases better than, the learning-based baselines on individual scales, and it attains superior overall results across the QMKP dataset. Notably, while NeuralQP, TriGNN, and GNNQP are specifically designed for quadratic problems, our approach targeting more general POIP problems with high-degree terms can still outperform these specialized models.

Table~\ref{tab:cflptc_comparison} shows the results on the quintic CFLPTC benchmark. The compared baselines include Gurobi and SCIP, as well as NeuralQP. Note that NeuralQP, TriGNN, and GNNQP were implemented only for quadratic POIP.
We compare with NeuralQP because it is a practically oriented method and shows strong performance on QMKP datasets. In contrast, GNNQP and TriGNN offer valuable insights into expressive power but are primarily theoretically oriented.
The model of NeuralQP was trained on a quadratic reformulation of CFLPTC obtained by introducing auxiliary variables and constraints, as detailed in Appendix B. Across all evaluated scales, our method produces significantly smaller primal gaps than the exact solvers and NeuralQP, demonstrating its ability to effectively exploit the quintic objective structure. Moreover, the performance of our method remains strong on the largest CFLPTC instances of $500\times 100$ scale, which are considerably larger than those seen during training, indicating good scalability and generalization across instance sizes.

\begin{table}[t!]
  \centering
  \setlength{\tabcolsep}{2pt}
  \fontsize{9}{11}\selectfont
  \renewcommand{\arraystretch}{0.8}
  \caption{Comparison on CFLPTC datasets in terms of mean and standard deviation of $\mathrm{gap_{\%}}$. The best results are highlighted in bold and $^{*}$ indicates the overall results are statistically different from the best one.}
  \begin{tabular}{cc|cccc|cccc}
    \toprule
    \multirow{2}{*}{Method} & \multirow{2}{*}{Train} & \multicolumn{4}{c}{Base solver: Gurobi} & \multicolumn{4}{c}{Base solver: SCIP} \\
    \cmidrule(lr){3-10}
    & & \begin{tabular}[c]{@{}l@{}}150$\times$\\30\end{tabular} & \begin{tabular}[c]{@{}l@{}}200$\times$\\30\end{tabular} & \begin{tabular}[c]{@{}l@{}}500$\times$\\100\end{tabular} & Overall & \begin{tabular}[c]{@{}l@{}}150$\times$\\30\end{tabular} & \begin{tabular}[c]{@{}l@{}}200$\times$\\30\end{tabular} & \begin{tabular}[c]{@{}l@{}}500$\times$\\100\end{tabular} & Overall \\
    \midrule
    \begin{tabular}[c]{@{}l@{}}Exact\\Solver\end{tabular} & -- & 51.42 & 58.20 & 38.66 & 48.89$_{\pm 11.11}^{*}$ 
    & 66.17 & 73.11 & 72.31 & 69.14$_{\pm 4.68}^{*}$ \\
    \midrule
    \multirow{2}{*}{\begin{tabular}[c]{@{}l@{}}Neural\\QP\end{tabular}} & \begin{tabular}[c]{@{}l@{}}50$\times$10\&\\50$\times$20\end{tabular} & 28.03 & 42.92 & 23.82 & \multirow{2}{*}{37.04$_{\pm 13.95}^{*}$}
    & 54.63 & 62.65 & 71.52 & \multirow{2}{*}{62.09$_{\pm 8.09}^{*}$} \\
    & \begin{tabular}[c]{@{}l@{}}150$\times$30\&\\200$\times$30\end{tabular} & 42.20 & 62.18 & 26.47 & 
    & 58.08 & 67.11 & 69.90 & \\
    \midrule
    \multirow{2}{*}{Ours} & \begin{tabular}[c]{@{}l@{}}50$\times$10\&\\50$\times$20\end{tabular} & 9.95 & 8.96 & 2.78 & \multirow{2}{*}{\textbf{6.59}$_{\pm 6.39}$}
    & 31.52 & \textbf{21.05} & 62.62 & \multirow{2}{*}{\textbf{42.01}$_{\pm 16.67}$} \\
    & \begin{tabular}[c]{@{}l@{}}150$\times$30\&\\200$\times$30\end{tabular} & \textbf{5.23} & \textbf{7.08} & \textbf{2.65} & 
    & \textbf{29.97} & 46.31 & \textbf{62.36} & \\
    \bottomrule
  \end{tabular}
  \label{tab:cflptc_comparison}
\end{table}

\subsection{Generalization to Other Problem Structures}
\label{sec: exp-generalization}

We next investigate how well the learned models generalize beyond the synthetic POIP benchmarks to problems with different structures. We consider the public QPLIB subset mentioned in Section~\ref{sec:exp-setup} and compare our method with all baselines. 
As shown in Table~\ref{tab:qplib_detailed_results}, our method attains the lowest (bold) or second lowest (italics) average gap on most instances, and the lowest overall average gap, demonstrating its capability to transfer well to structurally distinct problems.

We also investigate problems with nonlinear constraints. We used RandQCP datasets built by \cite{xiong2024neuralqp}, which are collections of quadratically constrained independent set problems. The nonlinearity in constraints is also represented by hyperedges connecting their variable vertices with other components in the hypergraph, while the HNN and repair-and-refinement process are unchanged. This indicates that our HNN has the potential to be easily extended to IP variants with polynomial terms in either objective functions or constraints.  The results, summarized in Table~\ref{QIS_results}, show that our method still achieves consistently superior performance compared to the baselines.

\begin{table}[t]
  \centering
  \caption{Comparison on QPLIB instances by mean value and standard deviation of $\mathrm{gap_{\%}}$. 
  The best and second best results are highlighted in bold and italics, respectively. $^{*}$ indicates statistically significant difference to the best result.}
  \centering
  \label{tab:qplib_detailed_results}
  \fontsize{9}{11}\selectfont
  \renewcommand{\arraystretch}{0.8}
  \begin{tabular}{cccccc}
    \toprule
        Instance ID & Gurobi & NeuralQP & GNNQP & TriGNN & Ours \\
    \midrule
        2315 & 59.75 & 15.92 & \textit{13.10} & \textbf{12.54} & 13.49 \\
        2733 & 1.38 & 0.28 & \textit{0.18} & \textbf{0.10} & 0.38 \\
        2957 & 2.11 & 2.07 & 1.88 & \textit{1.13} & \textbf{0.71} \\
        3347 & 0.52 & \textbf{0.20} & 9.28 & 5.93 & \textit{0.34} \\
        3402 & \textbf{2.80} & 7.10 & 4.95 & \textit{4.12} & 6.64 \\
        3584 & 66.75 & 15.04 & \textbf{7.75} & 15.59 & \textit{15.03} \\
        3752 & 14.09 & \textit{1.70} & 3.55 & 5.04 & \textbf{1.30} \\
        3841 & 26.69 & 9.90 & 8.57 & \textbf{0.53} & \textit{6.13} \\
        3860 & 48.95 & 15.92 & \textit{15.56} & 16.06 & \textbf{14.69} \\
        3883 & 8.25 & \textbf{0.36} & \textit{0.43} & 0.74 & 0.63 \\
        5962 & 16.89 & \textit{15.04} & 19.49 & 15.76 & \textbf{7.06} \\
    \midrule
        overall & 22.56$_{\pm 23.55}^{*}$ & 7.58$_{\pm 6.60}^{*}$ & 7.67$_{\pm 6.00}^{*}$ & 7.05$_{\pm 6.32}$ & \textbf{6.04}$_{\pm 5.70}$ \\
  \bottomrule
\end{tabular}
\end{table}

\begin{table}[t]
  \centering
  \caption{Results on RandQCP datasets by mean and standard deviation of $\mathrm{gap_{\%}}$.
  The best results are highlighted in bold and $^{*}$ indicates the overall results have statistically significant difference to the best one.}
  \setlength{\tabcolsep}{1pt}
  \fontsize{9}{11}\selectfont
  \renewcommand{\arraystretch}{0.8}
  \label{QIS_results}
  \begin{tabular}{cc|ccccc|ccccc}
    \toprule
       \multirow{2}{*}{Method} & \multirow{2}{*}{Train} & \multicolumn{5}{c|}{Base solver: Gurobi} & \multicolumn{5}{c}{Base solver: SCIP} \\
        \cmidrule(lr){3-12}
        & & 1k & 2k & 5k & 10k & Overall & 1k & 2k & 5k & 10k & Overall \\
        \midrule
        \begin{tabular}[c]{@{}l@{}}Exact\\solver\end{tabular} & -- & 3.21 & 6.45 & 5.83 & 9.88 & 6.34$_{\pm 9.84}^{*}$ 
        & 38.82 & 42.08 & 53.12 & 53.07 & 46.77$_{\pm 6.60}^{*}$ \\
        \midrule
        \multirow{3}{*}{\begin{tabular}[c]{@{}l@{}}Neural\\QP\end{tabular}} & Mini & 0.47 & 0.43 & 0.25 & 0.22 & \multirow{3}{*}{0.33$_{\pm 0.22}^{*}$} 
        & 0.60 & 0.51 & 0.37 & 0.31 & \multirow{3}{*}{0.43$_{\pm 0.23}^{*}$}\\
         & 1000 & 0.54 & 0.42 & 0.22 & 0.22 & 
         & 0.62 & 0.50 & 0.36 & 0.32 & \\
         & 2000 & -- & 0.39 & 0.25 & 0.19 & 
         & -- & 0.47 & 0.34 & 0.29 & \\
         \midrule
        \multirow{3}{*}{GNNQP} & Mini & 3.53 & 2.29 & 3.23 & 3.36 & \multirow{3}{*}{3.07$_{\pm 0.68}^{*}$} 
        & 3.60 & 2.40 & 3.29 & 3.42 & \multirow{3}{*}{3.14$_{\pm 0.66}^{*}$} \\
         & 1000 & 3.48 & 2.30 & 3.25 & 3.36 & 
         & 3.62 & 2.40 & 3.31 & 3.40 & \\
         & 2000 & -- & 2.32 & 3.23 & 3.38 & 
         & -- & 2.37 & 3.29 & 3.41 & \\
         \midrule
        \multirow{3}{*}{TriGNN} & Mini & 0.55 & 0.45 & 0.28 & 0.27 & \multirow{3}{*}{0.37$_{\pm 0.23}^{*}$} 
        & 0.67 & 0.49 & 0.37 & 0.29 & \multirow{3}{*}{0.48$_{\pm 0.88}^{*}$} \\
         & 1000 & 0.57 & 0.47 & 0.27 & 0.25 & 
         & 0.72 & 0.53 & 0.38 & 0.32 & \\
         & 2000 & -- & 0.50 & 0.28 & 0.23 & 
         & -- & 0.47 & 0.36 & 0.70 & \\
         \midrule
        \multirow{3}{*}{Ours} & Mini & \textbf{0.32} & \textbf{0.31} & \textbf{0.19} & \textbf{0.14} & \multirow{3}{*}{\textbf{0.26}$_{\pm 0.19}$} 
        & \textbf{0.46} & \textbf{0.37} & \textbf{0.27} & \textbf{0.17} & \multirow{3}{*}{\textbf{0.34}$_{\pm 0.19}$} \\
         & 1000 & 0.43 & 0.32 & 0.23 & 0.17 & 
         & 0.50 & 0.42 & 0.30 & 0.21 & \\
         & 2000 & -- & 0.38 & 0.20 & 0.18 & 
         & -- & 0.40 & 0.32 & 0.26 & \\
    \bottomrule 
  \end{tabular}
\end{table}

\subsection{Ablation Study}
\label{sec: exp-ablation}

We perform an ablation study on QMKP to evaluate the contribution of each architectural component. Two variants disable one of the convolution modules: \textbf{w/o-HyConv}, which removes the hyperedge-based convolution, and \textbf{w/o-VCConv}, which removes the variable–constraint-based convolution. For fairness, both variants use minimally modified graph representations that preserve all necessary relationships (Appendix C), while all other settings remain identical to our full model. To further examine the role of our hyperedge-based convolution in modeling high-order variable interactions, we introduce two additional variants that replace it with high-degree aggregation mechanisms from prior methods: \textbf{NeuralQP-HD}, which adopts NeuralQP’s high-degree handling block, and \textbf{GNNQP-HD}, which uses the corresponding module from GNNQP. In both cases, only the high-order processing component is altered.

All models are trained on QMKP-1k and evaluated on QMKP-1k and QMKP-2k, using Gurobi for repair-and-refinement and measuring solving quality via $\mathrm{gap_{\%}}$ under identical time limits. As shown in Table~\ref{tab:ablation}, removing either convolution module reduces performance, and substituting our high-order mechanism with that from NeuralQP or GNNQP also leads to worse results.

\begin{table}[htbp]
  \centering
  \caption{Comparison of our model and ablation baselines on QMKP instances in terms of $\mathrm{gap_{\%}}$. The best result of each testing dataset is highlighted in bold.}
  \fontsize{9}{11}\selectfont
  \renewcommand{\arraystretch}{0.8}
  \label{tab:ablation}
  \begin{tabular}{l|ccccc}
    \toprule
        & NeuralQP-HD & GNNQP-HD & w/o-VCConv & w/o-HyConv & Ours \\
    \midrule
        1k      & 0.33 & 0.36 & 0.42 & \textbf{0.30} & 0.31 \\
        2k      & 0.90 & 2.16 & 1.92 & 0.81 & \textbf{0.10} \\
        Average & 0.61 & 1.26 & 1.17 & 0.56 & \textbf{0.20} \\
    \bottomrule
  \end{tabular}
\end{table}

\section{Conclusion}
\label{sec: conclusion}
This paper introduces a novel hypergraph neural network (HNN) framework for polynomial-objective integer programming problems (POIP). Our approach contributes two key innovations: a high-degree-term-aware hypergraph representation that captures variable interactions in high-degree terms and variable-constraint interdependencies inherent in POIP problems, and a hypergraph neural network architecture that integrates hypergraph-based and bipartite-graph-based convolutions to enable accurate solution prediction. Comprehensive experimental evaluations across quadratic and quintic programming problems demonstrate that our method outperforms both state-of-the-art exact solvers and specialized learning-based approaches, establishing its effectiveness and practical value for challenging POIP applications.

While promising, our work represents just one step towards addressing the broader challenges of nonlinear integer programming. Future research directions include 1) designing more comprehensive representations and neural network structures for general nonlinear instances such as those with trigonometric and logarithmic functions and 2) exploring end-to-end frameworks that directly output feasible solutions without requiring repair mechanisms. 

\section*{Supplementary Materials}
Code and appendix are available at \url{https://github.com/PineappleLiMs/Solving_POIP_with_HNN.git}. In appendix, we provide more details of our HNN-based framework (Appendix~A), the benchmarks (Appendix~B), and our implementation (Appendix~C). We also report additional results, including the predictive performance (Appendix~D.1), CFLPTC comparisons between the native and quadratic formulations (Appendix~D.2), shifted geometric mean (Appendix~D.3), and comparisons across different solver random seeds (Appendix~D.4), to further substantiate the superiority of our HNN. Finally, Appendix~E presents the complexity analysis of our method.

\section{Acknowledgments.} This research is part of the LEO (Learning and Explaining Optimization) project which is co-funded by Holland High Tech | TKI HSTM via the PPS allowance scheme for public-private partnerships, and by Dassault Systèmes. This work used the Dutch national e-infrastructure with the support of the SURF Cooperative using grant no. EINF-11638.

%
%
%
\par\addvspace{1.5\baselineskip}
\begingroup
  \let\clearpage\relax
  \let\cleardoublepage\relax

  \bibliographystyle{splncs04}
  \bibliography{ref}
\endgroup





\appendix
\section{Details of the HNN-based Framework}
We present two key details of the HNN-based framework that were not covered in Section 4, allowing interested readers to reproduce our work.

\subsection{Raw Features of Hypergraph Representation}
\label{sec: raw_features}

We present the raw features of our hypergraph representation in Table \ref{Tab: raw_feature}. The table organizes four key components of our hypergraph representation that participate in convolutions. Each row corresponds to one component, with the first column identifying the component name, the second column listing its raw features, and the third column providing detailed descriptions of these features. Specifically, the variable vertices $\mathcal{V}$ are assigned nine-dimensional raw features that encode variable types, bound information, and their roles in the objective function. Constraint vertices $\mathcal{C}$ are assigned four-dimensional raw features based on their constraint sense and right-hand-side values. Hyperedges $\mathcal{H}$ are assigned raw features, the length of which varies according to the number of variables they contain, as introduced in Section 4.1. For a variable $v$ contained in a hyperedge $\epsilon$, a feature vector $\omega_{v\epsilon}$ containing the term coefficient and the variable's exponent is added to $\epsilon$'s raw features. Finally, standard edges $\mathcal{E}$ are assigned two-dimensional features that reflect coefficients and degrees of the corresponding variables within their associated constraints.

\begin{table*}[htbp]
\caption{Raw Features of High-Degree Term-Aware Hypergraph Representation}
\label{Tab: raw_feature}
\centering
\begin{tabular}{cp{2.2cm}p{8cm}}
\toprule
Tensor & Feature & Description \\
\midrule
\multirow{7}{*}{$\mathcal{V}$} & type & (continuous, binary, integer) as a one-hot encoding \\
& lb & Lower bound value of the variable \\
& up & Upper bound value of the variable \\
& inf\_lb & Binary indicator (1 if the lower bound is negative infinity, 0 otherwise) \\
& inf\_ub & Binary indicator (1 if the upper bound is positive infinity, 0 otherwise) \\
& avg\_obj\_coe & Average value of coefficients associated with this variable in the objective function \\
& avg\_obj\_deg & Average degree of this variable across all terms in the objective function \\
\midrule
\multirow{2}{*}{$\mathcal{C}$} & sense & ($<$, $>$, $=$) as a one-hot encoding \\
& rhs & Numerical value on the right-hand side of the constraint \\
\midrule
\multirow{2}{*}{$\omega_{v\epsilon}$} & deg & Degree of each variable in the high-degree term \\
& coe & Coefficient value associated with the high-degree term \\
\midrule
\multirow{2}{*}{$\mathcal{E}$} & avg\_coe & Average value of coefficients across all terms containing the variable in the associated constraint \\
& avg\_deg & Average degree of this variable across all terms containing it in the associated constraint \\
\bottomrule
\end{tabular}
\end{table*}

\subsection{Neighborhood Search for Repair-and-Refinement}
\label{sec: r&r_details}
We adopt parallel neighborhood optimization proposed by \cite{ye2023gnn,xiong2024neuralqp}, which incorporates two key components: a Q-repair-based repair strategy that efficiently converts model predictions to feasible solutions, and an iterated multi-neighborhood search that refines these solutions to achieve higher quality. In the following, we provide detailed descriptions of both components.

\subsubsection{Q-Repair-Based Repair Strategy}
\label{sec: exp_detail-repair}

The Q-repair begins by selecting the $\alpha n$ variables with the largest predicted loss values to optimize, while fixing the remaining $(1-\alpha)n$ variables to their predicted values. Here, $\alpha\in[0,1]$ is a proportion that determines the neighborhood search size, and $n$ represents the total number of variables. Then Q-repair traverses constraints to identify those that cannot be satisfied. This identification follows a greedy approach: calculating the upper and lower bounds of each term on the left-hand side, summing these bounds, and comparing the result with the right-hand side. When an unsatisfied constraint is detected, the variables involved in this constraint are incrementally added to the neighborhood until either all variables from that constraint have been incorporated or the neighborhood reaches a size limit of $\alpha_{\text{ub}}n$ variables. Q-repair terminates after evaluating all constraints and returns the neighborhood (i.e., variables to be optimized) for repair.

Subsequently, the repair strategy employs exact solvers (such as Gurobi and SCIP) to optimize the subproblem defined by the Q-repair neighborhood. If no feasible solution is identified within the allocated time, Q-repair is repeated with an enlarged initial $\alpha=\alpha_{\text{step}}+\text{len}(\text{neighborhood})/n$, followed by another neighborhood search on the new expanded neighborhood. This iterative process continues until a feasible solution is found, or $\alpha$ exceeds 1, or the maximum time to repair-and-refine has been reached.

\subsubsection{Iterated Multi-Neighborhood Search}

The iterated multi-neighborhood search begins by generating a set of initial neighborhoods using a sequential filling approach. Specifically, this process first randomly shuffles all constraints. Then, it iteratively processes each constraint by sequentially adding its variables to the current neighborhood. When the predefined neighborhood size limit is reached, a new neighborhood is created and the process continues, until all constraints and their associated variables have been assigned to neighborhoods. This process creates multiple neighborhoods where variables from the same constraint tend to appear together in the same neighborhoods, thereby reducing the likelihood of constraint violations. Next, using the solution obtained by the repair strategy as a starting point, subproblems are formulated based on each neighborhood and optimized using exact solvers.

After that, the algorithm generates crossover neighborhoods to explore combinations of different subproblem solutions. It groups all neighborhoods into pairs. For two neighborhoods $N_1$ and $N_2$ in a pair with their respective subproblem solutions $x^{(1)}, x^{(2)}$, assuming $x^{(1)}$ has an objective value equal to or better than $x^{(2)}$, a crossover neighborhood is created through two steps: 1) constructing a crossover solution $x^{\prime}$ by taking $x^{\prime}_i=x^{(1)}_i$ for variables in $N_1$ and $x^{\prime}_i=x^{(2)}_i$ for other variables, and 2) applying Q-repair to $x^{\prime}$. Then, subproblems based on these crossover neighborhoods are optimized. The algorithm selects the best solution among all the candidates, both initial neighborhoods and crossover neighborhoods, to serve as the starting point for the next iteration. These two processes repeat until the predetermined time limit is reached, and the best solution found across all iterations is returned as the final result.

\section{Details of Benchmarks}
\label{sec: benchmark_detail}
This section introduces the details of the synthetic datasets used in our experiments.
\subsection{Details of Synthetic Quadratic Instances}
In Section 5.2 and Section 5.3, we evaluate the efficiency of our HNN-based framework using two synthetic quadratic datasets: QMKP and RandQCP, which are generated and provided by \cite{xiong2024neuralqp}. The formulations of these problems are presented below.

The Quadratic Multiple Knapsack Problem (QMKP) extends the classic knapsack problem by incorporating multiple weight constraints and quadratic profit terms. It involves selecting items to place in a knapsack with limited capacity across multiple weight dimensions. Each item yields an individual profit, while specific pairs of items generate additional interactive profits when selected together. The objective is to maximize the total profit while adhering to capacity constraints. QMKP can be formulated as a quadratic programming problem as shown in Eqs.~\ref{eq: qmkp-obj}-\ref{eq: qmkp-var}:
\begin{align}
\max \quad & \sum_{i} c_i x_i + \sum_{(i,j) \in E} q_{ij} x_i x_j, \label{eq: qmkp-obj}\\
\text{s.t.} \quad & a_i^k x_i \leq b^k, \quad \forall k \in M, \label{eq: qmkp-con1}\\
& x_i \in \{0,1\}, \quad \forall i \in N, \label{eq: qmkp-var}
\end{align}
where $x_i$ is a binary variable indicating whether item $i$ is selected, $c_i$ represents the individual profit for item $i$, and $q_{ij}$ denotes the interactive profit obtained by selecting both items $i$ and $j$. The set $E$ contains item pairs with interactive profits, $a_i^k$ represents the $k$-th weight of item $i$, and $b^k$ denotes the knapsack's capacity on the $k$-th weight dimension. $M$ and $N$ represent the total number of weight dimensions and items, respectively.

The Random Quadratically Constrained Quadratic Program (RandQCP) is an extension of the independent set problem. It aims to select vertices from a hypergraph to maximize total weights while satisfying specified constraints on each hyperedge. The quadratic programming formulation of RandQCP is given in Eqs.~\ref{eq: randqcp-obj}-\ref{eq: randqcp-var}.
\begin{align}
\max \quad & \sum_{i \in V} c_i x_i, \label{eq: randqcp-obj}\\
\text{s.t.} \quad & \sum_{i \in e} a_i x_i + \sum_{i,j \in e, i \neq j} q_{ij}x_i x_j - |e| \leq 0, \quad \forall e \in \mathcal{E}, \label{eq: randqcp-con1}\\
& x_i \in \{0,1\}, \quad \forall i \in V, \label{eq: randqcp-var}
\end{align}
where $V$ represents the set of vertices, $\mathcal{E}$ denotes the hyperedge set, $c_i$ is the weight associated with vertex $i$, and $a_i$ and $q_{ij}$ are the limitation coefficients for selecting vertex $i$ and vertex pair $(i,j)$, respectively. The term $e$ refers to a specific hyperedge, and $|e|$ indicates the number of vertices contained within hyperedge $e$.

For details of generation and access to the generated datasets, please refer to \cite{xiong2024neuralqp}.

\subsection{Details of Synthetic Quintic Instances}
To evaluate the effectiveness of our HNN-based method on more complex integer programming problems, we generated synthetic quintic datasets based on the Capacitated Facility Location Problem under Traffic Congestion (CFLPTC) inspired by \cite{bai2011biofuel} and \cite{holmberg1999exact}. The formulation and generation procedures are detailed below.

\subsubsection{Formulation of CFLPTC}

CFLPTC extends the standard capacitated facility location problem by incorporating traffic congestion effects. Consider a scenario with $m$ customers $J=\{1,\cdots,m\}$ and $n$ potential facility locations $I=\{1,\cdots,n\}$. Each customer $j$ has a demand $D_j$, while each facility at location $i$ incurs an opening cost $o_i$ and has a capacity $C_i$. Once opened, a facility can serve customers provided that the total demand it satisfies does not exceed its capacity. Each customer must be served by exactly one opened facility. The transportation cost for serving customer $j$ from facility $i$ depends on the distance between them $d_{ij}$ and the traffic congestion level. The objective is to determine which facilities to open and how to assign customers to these facilities, so that the total cost comprising facility opening costs and transportation expenses is minimized. The mathematical formulation is presented in Eqs.~\ref{eq: cflptc-obj}-\ref{eq: cflptc-var}.

\begin{align}
\max \quad & -\sum_{i\in I}o_i y_i - \sum_{i\in I}\sum_{j\in J}\alpha(1+0.15e_i^\beta)d_{ij}x_{ij} \label{eq: cflptc-obj} \\
\text{s.t.} \quad & \sum_{i}x_{ij}=1, \forall j \in J, \label{eq: cflptc-con1} \\
& x_{ij}\le y_i, \forall i\in I, j\in J, \label{eq: cflptc-con2}\\
& \sum_{j}D_j x_{ij} \le C_i y_i, \forall i \in I, \label{eq: cflptc-con3}\\
& e_i = \frac{\sum_{j}D_j x_{ij}+b_i}{T_i}, \forall i \in I, \label{eq: cflptc-con4}\\
& x_{ij},y_i\in \{0,1\}, \forall i\in I, j\in J. \label{eq: cflptc-var}
\end{align}
where $y_i$ and $x_{ij}$ are binary variables to determine whether to open the facility at location $i$ and whether to assign customer $j$ to the facility at location $i$, separately. 

In the objective function in Eq.~\ref{eq: cflptc-obj}, the transportation cost from facility $i$ to customer $j$ is expressed as $\alpha(1+0.15e_i^\beta)d_{ij}x_{ij}$, where the term $\alpha(1+0.15e_i^\beta)$ quantifies the additional cost induced by traffic congestion. This formulation, together with Eq.~\ref{eq: cflptc-con4} which determines $e_i$, is derived from the Bureau of Public Roads (BPR) function, an empirical formula for estimating increased transportation time corresponding to congestion level \cite{united1964traffic}. In this context, $T_i$ represents the total traffic capacity surrounding facility location $i$ and $b_i$ denotes the background traffic flow in the vicinity. The parameters $\alpha$ and $\beta$ are typically set to 1 and 4, respectively, which makes CFLPTC a quintic programming problem.

While CFLPTC technically falls under the category of mixed-integer programming due to its combination of binary variables ($x_{ij}$, $y_i$) and continuous variables ($e_i$), it remains essentially an integer programming problem. This is because the continuous variables $e_i$ are merely auxiliary and are completely determined by the binary assignment variables $x_{ij}$. Therefore, it is methodologically reasonable to include CFLPTC as a dataset in this work, which focuses on integer programming problems.

\subsubsection{Quadratic Reformulation of CFLPTC}

In Section 5.1, we compared our method against NeuralQP on the quintic CFLPTC instances. However, NeuralQP is designed exclusively for quadratic optimization problems and cannot directly handle the quintic terms present in the original CFLPTC formulation. To enable this comparison, we reformulated the quintic CFLPTC instances into equivalent quadratic problems by introducing auxiliary variables that decompose higher-order terms. The reformulation strategy systematically replaces quintic terms with chains of quadratic relationships. Specifically, for each $i\in I$, we define $e_{1i} = e_i^2$ and $e_{2i} = e_{1i}^2$, which transform the quintic terms $e_i^4 x_{ij}$ into quadratic terms $e_{2i}x_{ij}$. The complete quadratic reformulation is presented in Eq.~\ref{eq: cflptc-obj2}-\ref{eq: auxiliary2}.

\begin{align}
\max \quad & -\sum_{i\in I}o_i y_i - \sum_{i\in I}\sum_{j\in J}\alpha(1+0.15e_{2i})d_{ij}x_{ij} \label{eq: cflptc-obj2} \\
\text{s.t.} \quad& \text{constraints~\ref{eq: cflptc-con1} - \ref{eq: cflptc-var}}, \\
& e_{1i} = e_i^2, \forall i \in I, \label{eq: auxiliary1}\\
& e_{2i} = e_{1i}^2, \forall i \in I, \label{eq: auxiliary2}
\end{align}

It is important to note that while lower-degree objective functions and constraints are generally more tractable for optimization algorithms than their higher-degree counterparts, the reformulation process inevitably introduces additional variables and constraints that can impose significant computational overhead. For CFLPTC instances, the quadratic reformulation requires $2n$ additional variables ($e_{1i}, e_{2i}$) and $2n$ additional quadratic constraints (Eq.~\ref{eq: auxiliary1} and \ref{eq: auxiliary2}), substantially increasing the complexity. The increase of complexity may offset or even outweigh the computational benefits gained from degree reduction, as solvers must now handle a larger search space and a more complicated constraint set. Consequently, reformulating high-degree problems into lower-degree equivalents does not guarantee improved optimization efficiency; the net effect depends on the trade-off between reduced degree and increased problem complexity, which varies with specific problem characteristics and solver capabilities. This trade-off underscores the importance of developing optimization methods that can directly handle high-degree integer programming problems rather than relying solely on quadratic reformulations.

\subsubsection{Instance Generation}

Following the approach in \cite{holmberg1999exact}, we generated datasets at four distinct scales for training, as detailed in Table \ref{tab: generating_information}. The notation $U(a,b)$ indicates that the corresponding parameters are randomly sampled from a uniform distribution ranging from $a$ to $b$ (inclusive). Both customer and facility locations were generated within a two-dimensional Euclidean space according to the "Coordinate" specifications in Table \ref{tab: generating_information}, with distances calculated using the Euclidean metric. Consistently across all datasets, the total traffic capacity $T_i$ was generated as $U(1,4) \cdot C_i$, while the background traffic flow $b_i$ was set to $U(0.1,1) \cdot T_i$.

\begin{table*}[htbp]
\caption{Setting for CFLPTC Training Dataset Generation}
  \label{tab: generating_information}
  \centering
  \begin{tabular}{cccccccc}
    \toprule
    Dataset & Number & $m$ & $n$ & Coordinate & $D_j$ & $o_i$ & $C_i$ \\
    \midrule
    1 & 1605 & 50 & 10 & $U(10, 200)$ & $U(10, 50)$ & $U(300, 700)$ & $U(100,500)$ \\
    2 & 1119 & 50 & 20 & $U(10, 200)$ & $U(30, 80)$ & $U(300, 700)$ & $U(100,500)$ \\
    3 & 984 & 150 & 30 & $U(10, 300)$ & $U(10, 50)$ & $U(300, 700)$ & $U(200,600)$ \\
    4 & 200 & 200 & 30 & $U(10, 200)$ & $U(10, 50)$ & $U(500,1500)$ & $U(500,800)$ \\
    \bottomrule
  \end{tabular}
\end{table*}

For testing purposes, we generated 16 instances each at the $150 \times 30$ scale and the $200\times 30$ scale, adhering to the same parameter settings used for training datasets 3 and 4, respectively. Additionally, we created 10 larger instances at the $500 \times 100$ scale, following the parameter settings of training dataset 1 but with adjusted values for $m$ and $n$. These testing datasets enable a comprehensive evaluation of our model's capability to effectively tackle complex, large-scale integer programming problems with high-degree terms.

\section{Implementation Details}
\label{sec: config}
\paragraph{Model Details}
First, all raw features of the input hypergraph were transformed into initial embeddings through 2-layer MLPs activated by LeakyReLU, where the dimensions of hidden spaces and output features are 64 and 16, respectively. The number of iterations for executing hyperedge-based convolution is $L=6$. The negative slopes of all LeakyReLU activations are set to 0.1.

\paragraph{Training Details}
We utilized AdamW with a learning rate of 1e-4 and weight decay of 1e-4 as the optimizer to train our model. We set the batch size to 64 and training epochs to 100. On each training dataset, our HNN models were trained on a supercomputer node with an NVIDIA A100 GPU and an 18-core Intel Xeon Platinum 8360Y CPU. For fair comparison, we used the same device to train the models of learning-based baselines, with the same hyper-parameter settings as in their original papers. 

\paragraph{Inference Details}
We used Gurobi 12.0.0 and SCIP 9.2.0 for all inference tests, which were run exclusively on CPUs. Tests using SCIP were conducted on a supercomputer equipped with an AMD Rome 7H12 CPU, while those using Gurobi were run on a separate supercomputer with an Intel Xeon Platinum 8260 CPU. Note that this setup does not introduce unfairness, as our comparisons focus on the performance of different methods within the same exact solver, rather than comparing the solvers themselves.

We implemented the repair-and-refinement algorithm (see Appendix~\ref{sec: r&r_details}) following the parameter settings proposed by \cite{xiong2024neuralqp}. Specifically, for the Q-repair-based repair strategy, we initialized the parameter $\alpha$ at 0.1, with $\alpha_{\text{ub}} = 1$ and $\alpha_{\text{step}} = 0.05$. For the iterated multi-neighborhood search, the neighborhood size is defined as half the number of problem variables. For each subproblem occurring in both the Q-repair-based repair strategy and the iterated multi-neighborhood search, we set a maximum wall-clock time of 60 seconds when addressing largest-scale instances: 10k-scale QMKP and RandQCP problems, and 500×100-scale CFLPTC datasets. All other testing datasets were limited to 30 seconds per subproblem. The repair-and-refinement stops when the total wall-clock time reaches the preset limit (see Section 5.1).

\paragraph{Details of the Ablation Baselines}
In the ablation studies (Section 5.4), we constructed two ablation baselines (w/o-HyConv and w/o-VCConv) to investigate the contributions of hyperedge-based convolution and variable-constraint-based convolution, as well as two additional ablation baselines (NeuralQP-HD and GNNQP-HD) to examine the role of our hyperedge-based convolution in parsing high-order relationships from high-degree terms. The first two ablation baselines are constructed to be as comparable as possible to our HNN model while omitting the targeted convolution modules. Since simply removing a component would prevent the model from capturing one key relationship in POIP, we make slight but necessary adjustments to their input representations. For w/o-HyConv, the only change is the removal of hyperedges from the representation. For w/o-VCConv, its hypergraph representation contains the same variable and constraint vertices as in our representation but differs in that it has no edges and uses alternative hyperedges. These hyperedges encode both variable interactions in high-degree terms and variable-constraint interdependencies: each term is represented by a hyperedge connecting its variables and the constraint to which it belongs. The hyperedge features follow the same design as our representation.

For NeuralQP-HD and GNNQP-HD, we replace the hyperedge-based convolution of our model with the convolutions for high-degree terms from NeuralQP~\cite{xiong2024neuralqp} and GNNQP~\cite{chen2025expressive}, respectively. Specifically, NeuralQP-HD adds additional vertices to its hypergraph representation to represent degrees, and each of its hyperedges connects variable vertices and degree vertices if one high-degree term contains the corresponding variables with the exponents of the corresponding degrees. It also applies a two-step convolution similar to our Eq. 5 and Eq. 6 to capture high-order relationships. GNNQP-HD uses hyperedges to connect variable vertices if they appear in the same high-degree term, and it lets a variable vertex appear $k$ times in the hyperedge if its variable's degree is $k$ in this term. To pass information from high-order relationships, it also uses convolution layers to aggregate hyperedges into variable embeddings. Other model structures, including the embedding initialization, the variable-constraint convolution, and the output layer, all remain the same as our model for fairness.

\section{Additional Experiments to Evaluate Model Prediction}
\label{sec: exp_additional}

\subsection{Predictive Performance}
\label{sec: exp_additional-prediction}

This section investigates the proposed HNN's predictive performance. We applied our HNN models trained on QMKP 1k-scaled training data to the QMKP test sets with 1k-scaled instances. We use NeuralQP as the comparison baseline. The prediction performance is examined in terms of the relative primal gap in percentage ($\mathrm{gap}_{\%}$, see Section 5.1) on the feasible solutions that are converted from the models' predictions via the Q-Repair-Based Repair Strategy based on Gurobi (detailed in Appendix A.2). No further refinements are performed on these feasible solutions.

The results in Table \ref{tab:predict_comparison} show that the feasible solutions repaired from our model's predictions have a closer gap to the best-known solutions, indicating better solution quality. These results demonstrate our model's superior predictive capability.

\begin{table*}[h]
\caption{Comparison on QMKP-1k datasets in terms of prediction performance.}
\centering
\label{tab:predict_comparison}
\begin{tabular}{c|cc}
\toprule
Method & NeuralQP & Ours \\
\midrule
$\mathrm{gap}_{\%}$ & 99.10 & \textbf{71.92} \\
\bottomrule
\end{tabular}
\end{table*}

\subsection{Comparison on CFLPTC: Native Quintic vs. Quadratic Reformulation}
\label{sec: exp_additional-Quad}

To assess whether the performance gains reported in Table~2 are primarily due to leveraging the native quintic objective structure, we additionally trained our method on the quadratic reformulation of CFLPTC (see Appendix~B.2). We keep the HNN architecture and training protocol unchanged, yielding a like-for-like variant (“Ours–Quad”). As shown in Table~\ref{tab:cflptc_comparison2}, Ours-Quad outperforms the other baselines but consistently under-performs our default model trained on the native quintic formulation, indicating that our approach benefits from explicitly leveraging the higher-order objective structure.

\begin{table}[t!]
  \centering
  \setlength{\tabcolsep}{2pt}
  \fontsize{9}{11}\selectfont
  \renewcommand{\arraystretch}{0.8}
  \caption{Comparison on CFLPTC datasets in terms of mean and standard deviation of $\mathrm{gap_{\%}}$. The best results are highlighted in bold and $^{*}$ indicates the overall results are statistically different from the best one.}
  \begin{tabular}{cc|cccc}
    \toprule
    \multirow{2}{*}{Method} & \multirow{2}{*}{Train} & \multicolumn{4}{c}{Base solver: Gurobi}\\
    \cmidrule{3-6}
    & & \begin{tabular}[c]{@{}l@{}}150$\times$\\30\end{tabular} & \begin{tabular}[c]{@{}l@{}}200$\times$\\30\end{tabular} & \begin{tabular}[c]{@{}l@{}}500$\times$\\100\end{tabular} & Overall \\
    \midrule
    \begin{tabular}[c]{@{}l@{}}Exact\\Solver\end{tabular} & -- & 51.42 & 58.20 & 38.66 & 48.89$_{\pm 11.11}^{*}$ \\
    \midrule
    \multirow{2}{*}{\begin{tabular}[c]{@{}l@{}}Neural\\QP\end{tabular}} & \begin{tabular}[c]{@{}l@{}}50$\times$10\&\\50$\times$20\end{tabular} & 28.03 & 42.92 & 23.82 & \multirow{2}{*}{37.04$_{\pm 13.95}^{*}$} \\
    & \begin{tabular}[c]{@{}l@{}}150$\times$30\&\\200$\times$30\end{tabular} & 42.20 & 62.18 & 26.47 & 
    \\
    \midrule
    \multirow{2}{*}{\begin{tabular}[c]{@{}l@{}}Ours\\–Quad\end{tabular}} & \begin{tabular}[c]{@{}l@{}}50$\times$10\&\\50$\times$20\end{tabular} & 32.49 & 38.40 & 11.25 & \multirow{2}{*}{29.77$_{\pm 25.23}^{*}$} \\
    & \begin{tabular}[c]{@{}l@{}}150$\times$30\&\\200$\times$30\end{tabular} & 31.45 & 39.77 & 11.46 & 
    \\
    \midrule
    \multirow{2}{*}{Ours} & \begin{tabular}[c]{@{}l@{}}50$\times$10\&\\50$\times$20\end{tabular} & 9.95 & 8.96 & 2.78 & \multirow{2}{*}{\textbf{6.59}$_{\pm 6.39}$} \\
    & \begin{tabular}[c]{@{}l@{}}150$\times$30\&\\200$\times$30\end{tabular} & \textbf{5.23} & \textbf{7.08} & \textbf{2.65} & \\
    \bottomrule
  \end{tabular}
  \label{tab:cflptc_comparison2}
\end{table}

\subsection{Shifted Geometric Mean of $\mathrm{gap_{\%}}$}
\label{sec: exp_additional-sgm}
We also report the shifted geometric mean (SGM) of $\mathrm{gap_{\%}}$ using shift scale of 1, which is a commonly used metric for aggregating ratio-based performance. The results are shown in Table~\ref{tab:qmkp_sgm_gap}, where our method still achieves the best performance under this SGM measure.

\begin{table}[t]
  \caption{Comparison on QMKP datasets in terms of SGM (shift=1) of $\mathrm{gap_{\%}}$. The best results are highlighted in bold.}
  \label{tab:qmkp_sgm_gap}
  \centering
  \small
  \setlength{\tabcolsep}{6pt}
  \begin{tabular}{ccc}
    \toprule
    \multirow{2}{*}{Methods} & \multicolumn{2}{c}{Base solver} \\
     & Gurobi & SCIP \\
    \midrule
    Exact solver & 5.12 & 13.61 \\
    NeuralQP     & 0.10 & 4.95  \\
    GNNQP        & 1.18 & 5.94  \\
    TriGNN       & 0.12 & 5.03  \\
    Ours         & \textbf{0.08} & \textbf{4.75} \\
    \bottomrule
  \end{tabular}
\end{table}

\subsection{Robustness to Solver Random Seeds}
\label{sec: exp_additional-seeds}
To evaluate the robustness of our method’s performance gains, we conducted additional experiments with different random seeds for the backend solver. Specifically, we used Gurobi as the backend and compared our method against standalone Gurobi and NeuralQP on QMKP instances. For each method, we ran five independent trials, setting Gurobi’s random seed to 2, 10, 22, 24, and 60, respectively. As shown in Table~\ref{tab: QKP_results_seeded}, our method consistently delivers superior performance, matching the conclusions in Section~5.

\begin{table}[t]
  \centering
  \caption{Comparison of our method, NeuralQP and Gurobi under multiple solver random seeds on QMKP datasets in terms of mean $\mathrm{gap_{\%}}$. The best results are highlighted in bold.}
  \label{tab: QKP_results_seeded}
  \setlength{\tabcolsep}{1pt}
  \fontsize{9}{11}\selectfont
  \renewcommand{\arraystretch}{0.8}
  \begin{tabular}{cc|ccccc}
    \toprule
        Method & Train & 1k & 2k & 5k & 10k & Overall \\
        \midrule
        Gurobi & -- & $0.45$ & $0.22$ & $28.82$ & $24.89$ & $13.59$ \\
        \midrule
        \multirow{3}{*}{\begin{tabular}[c]{@{}l@{}}Neural\\QP\end{tabular}} & Mini & $0.35$ & $0.08$ & $0.04$ & $0.03$ & \multirow{3}{*}{0.11} \\
         & 1k & $0.36$ & $0.09$ & \textbf{0.03} & $0.04$ & \\
         & 2k & -- & $0.09$ & $0.05$ & $0.03$ & \\
        \midrule
        \multirow{3}{*}{Ours} & Mini & \textbf{0.17} & 0.10 & $0.03$  & 0.04 & \multirow{3}{*}{\textbf{0.09}} \\
         & 1k & $0.31$ & $0.10$ & $0.05$  & $0.04$  & \\
         & 2k & -- & \textbf{0.08} & $0.05$  & \textbf{0.03}  & \\
    \bottomrule
  \end{tabular}
\end{table}


\section{Complexity Analysis}

This section analyzes the memory requirements of the proposed high-degree-term-aware hypergraph representation and the arithmetic time complexity of the proposed HNN's inference. We consider a POIP instance with $n$ variables, $m$ constraints, and $n_h$ high-degree terms. Let $s$ denote the total number of variable occurrences across all high-degree terms, and let $n_e$ denote the total number of variable-constraint incidences (i.e., the number of times any variable appears with a nonzero coefficient in any constraint). We estimate the efficiency of our method in terms of both memory usage and computational complexity in the following subsections.

\subsection{Memory Requirement for the High-Degree-Term-Aware Hypergraph Representation}
According to Section 4.1 and Appendix A.1, the hypergraph representation of the POIP instance comprises four components:
\begin{itemize}
    \item $n$ variable vertices, each with 9 raw features;
    \item $m$ constraint vertices, each with 4 features;
    \item $n_h$ hyperedges, with $s$ vertex-hyperedge coefficients, where each coefficient contains 2 floats;
    \item $n_e$ edges, each with 2 features;
\end{itemize}

Variable vertices and constraint vertices can be stored using their indices, while hyperedges and edges can be stored using tuples of vertex indices they contain. In total, hypergraph structure requires $(n + m + s + 2n_e)$ indices to represent. Additionally, there are $(9n + 4m + 2n_e + 2s)$ raw features. Assuming all indices are stored as 4-byte integers and raw features are stored as 8-byte floats (double precision), the total memory requirement for the hypergraph representation is:
\begin{equation} \label{eq: bytes}
    \textbf{bytes} = 76n + 36m + 20s + 24n_e.
\end{equation}

To illustrate this with a concrete example, consider the largest CFLPTC instances we tested, which involve 500 customers and 100 facilities. As detailed in Section B.2, these instances have $n = 50200, m = 50700, n_e = 200300, s = 100000$. Applying Eq.~\ref{eq: bytes}, the total memory requirement is 12,447,600 bytes, or approximately 11.87 megabytes (MB). This represents a very manageable memory overhead for modern hardware, demonstrating that our hypergraph representation remains practical even for large-scale instances.

\subsection{Arithmetic Time Complexity for the HNN}

In this subsection, we analyze the arithmetic complexity of our HNN model during inference. Let $n_{\text{hid}}$ denote the largest dimension among raw features, hidden embeddings, and outputs, and assume we perform $L_{\text{hyper}}$ hypergraph-based convolutions and $L_{\text{bi}}$ bipartite-graph-based convolutions. The complexity analysis for each component is as follows:
\begin{itemize}
    \item Initial embedding: it is a 2-layer MLP applied on all raw features, with arithmetic complexity $O((n+m+s+n_e)n_{\text{hid}}^2)$;
    \item Hypergraph-based convolution: 
        \begin{itemize}
            \item Eq. 5 performs weighted summation with complexity $O(s n_{\text{hid}})$;
            \item Eq. 6 combines weighted means, a 2-layer MLP, and a residual connection, with complexity $O(sn_{\text{hid}})$, $O(nn_{\text{hid}}^2)$, and $O(nn_{\text{hid}})$, respectively. The total complexity is $O(sn_{\text{hid}}+nn_{\text{hid}}^2)$;
            \item Overall complexity: $O(L_{\text{hyper}}(sn_{\text{hid}}+nn_{\text{hid}}^2))$;
        \end{itemize}
    \item Bipartite-graph-based convolution:
        \begin{itemize}
            \item Eq. 7 combines summations, a 2-layer MLP, and residual connection, with complexity $O(n_e n_{\text{hid}}$, $O(mn_{\text{hid}}^2)$, and $O(mn_{\text{hid}})$, separately. The total complexity is $O(n_e n_{\text{hid}}+mn_{\text{hid}}^2)$;
            \item Eq. 8 has similar structure to Eq. 7, with complexity $O(n_e n_{\text{hid}}+nn_{\text{hid}}^2)$;
            \item Overall complexity: $O(L_{\text{bi}}(n_e n_{\text{hid}}+mn_{\text{hid}}^2+nn_{\text{hid}}^2))$;
        \end{itemize}
    \item Output layer: A 2-layer MLP applied to variable embeddings, with complexity $O(nn_{\text{hid}}^2)$.
\end{itemize}

Therefore, the overall arithmetic complexity of HNN inference is $O(n_{\text{hid}}(L_{\text{hyper}}s+L_{\text{bi}}n_e) + n_{\text{hid}}^2(L_{\text{hyper}}n+L_{\text{bi}}n+L_{\text{bi}}m))$. Since $n_{\text{hid}}$, $L_{\text{hyper}}$, and $L_{\text{bi}}$ are fixed constants in our experiments (see Appendix C), the arithmetic complexity simplifies to $O(n+m+s+n_e)$, which scales linearly with the number of variables, constraints, hyperedge density, and edge density.

To demonstrate robustness, we consider the extreme case of a fully dense hypergraph representation where every pair of variable and constraint vertices is connected by edges, and all variable vertices are connected within each hyperedge. In this scenario, $s=n_h n$ and $n_e=nm$ yield a quadratic complexity $O(n(m+n_e))$. This analysis shows that even in such extreme cases, our HNN model maintains acceptable computational efficiency for inference. Hypergraph representations for integer programming problems are typically sparse in both hyperedges and edges, making our HNN model highly efficient.

\end{document}